%% file: main.tex
\documentclass[10pt,twocolumn,letterpaper]{article}

\usepackage[pagenumbers]{cvpr} 

\usepackage{kotex} 
\usepackage[export]{adjustbox}            
\usepackage[flushleft]{threeparttable}    
\usepackage{booktabs}                     
\newcommand\mc[1]{\multicolumn{1}{c}{#1}} 
\usepackage{colortbl}

\usepackage{makecell}                     
\usepackage{comment}
\usepackage{graphicx} 
\usepackage{bm}
\usepackage{multirow} 

\usepackage{ifthen}
\newboolean{togglecolor}    
\setboolean{togglecolor}{false} 
\newcommand{\tcolor}[1]{%
    \ifthenelse{\boolean{togglecolor}}{\textcolor{blue}{#1}}{#1}%
}


\definecolor{cvprblue}{rgb}{0.21,0.49,0.74}
\usepackage[pagebackref,breaklinks,colorlinks,allcolors=cvprblue]{hyperref}

\def\paperID{*****} 
\def\confName{CVPR}
\def\confYear{2025}

\title{PIDLoc: Cross-View Pose Optimization Network Inspired by PID Controllers}

\author{
    Wooju Lee$^{1}$ Juhye Park$^{1}$ Dasol Hong$^{1}$ Changki Sung$^{1}$ Youngwoo Seo$^{2}$ Dongwan Kang$^{2}$ Hyun Myung$^{1}$ \\
$^{1}$Urban Robotics Lab, School of Elctrical Engineering, KAIST \qquad $^{2}$ Hanwha Aerospace   \\
        {\tt\small $^{1}$\{dnwn24,jhpark12,ds.hong,cs1032,hmyung\}@kaist.ac.kr $^{2}$\{youngwoo.seo,dongwan.kang\}@hanwha.com
        }
    }

\begin{document}
\maketitle
\input{sec/0_abstract}    
\input{sec/1_intro}
\input{sec/2_formatting}

\input{sec/3_finalcopy}
\input{sec/4_experiments}
{
    \small
    \bibliographystyle{ieeenat_fullname}
    \bibliography{main}
}


\end{document}

%% file: sec/0_abstract.tex
\newcommand{\dnnpose}[1]{SPE}
\begin{abstract}
Accurate localization is essential for autonomous driving, but GNSS-based methods struggle in challenging environments such as urban canyons. Cross-view pose optimization offers an effective solution by directly estimating vehicle pose using satellite-view images. However, existing methods primarily rely on cross-view features at a given pose, neglecting fine-grained contexts for precision and global contexts for robustness against large initial pose errors. To overcome these limitations, we propose PIDLoc, a novel cross-view pose optimization approach inspired by the proportional-integral-derivative (PID) controller. \tcolor{Using RGB images and LiDAR,} the PIDLoc comprises the PID branches to model cross-view feature relationships and the spatially aware pose estimator (\dnnpose{}) to estimate the pose from these relationships. The PID branches leverage feature differences for local context (P), aggregated feature differences for global context (I), and gradients of feature differences for precise pose adjustment (D) to enhance localization accuracy under large initial pose errors. Integrated with the PID branches, the \dnnpose{} captures spatial relationships within the PID-branch features for consistent localization. Experimental results demonstrate that the PIDLoc achieves state-of-the-art performance in cross-view pose estimation for the KITTI dataset, reducing position error by $37.8\%$ compared with the previous state-of-the-art.
\end{abstract}

%% file: sec/1_intro.tex
\section{Introduction}
\label{sec:intro}
Localization~\cite{lee20232, choi2020brm, shi2020looking, shi2022beyond, wang2023view, wang2024view} is a critical component of autonomous driving systems, enabling resilient navigation across diverse environments. While the global navigation satellite system (GNSS) is commonly used for geographic positioning, it is vulnerable to interference, jamming, and signal blockage in challenging environments such as urban canyons~\cite{reid2019localization, zidan2020gnss, pirayesh2022jamming}. Thus, autonomous driving systems relying on GNSS encounter significant challenges in GNSS-denied or GNSS-limited scenarios.

\begin{figure}[h]
    \centering
    \begin{subfigure}{0.15\textwidth}
        \centering
        \includegraphics[width=\textwidth]{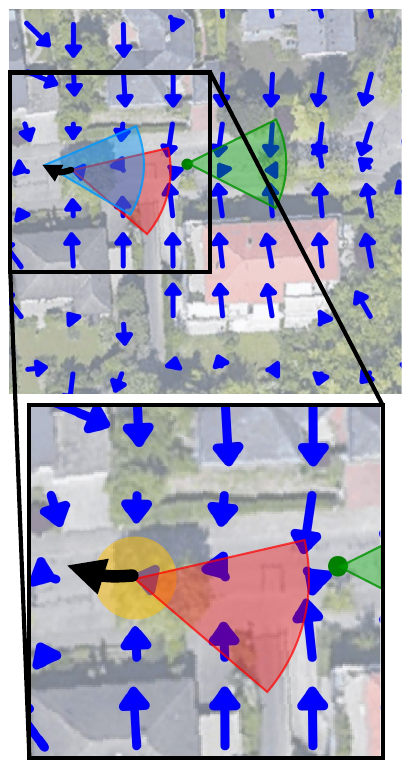}
        \caption{P branch}
        \label{fig:figure1_p}
    \end{subfigure}
    \begin{subfigure}{0.15\textwidth}
        \centering
        \includegraphics[width=\textwidth]{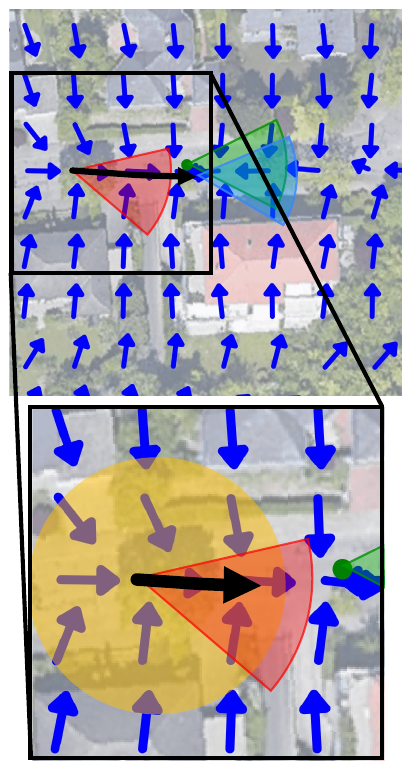}
        \caption{I branch}
        \label{fig:figure1_i}
    \end{subfigure}
    \begin{subfigure}{0.15\textwidth}
        \centering
        \includegraphics[width=\textwidth]{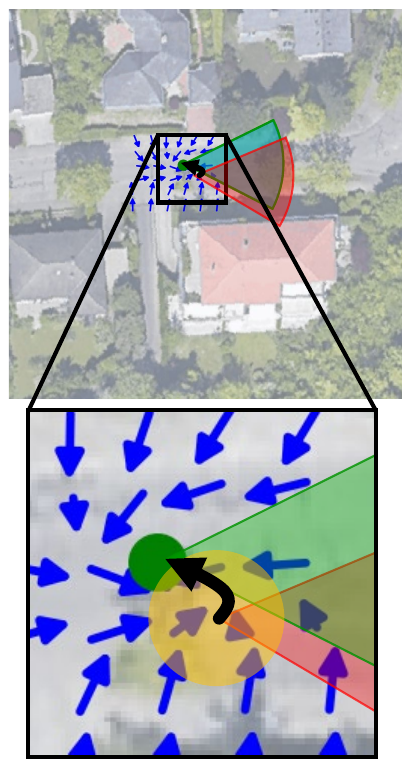}
        \caption{D branch}
        \label{fig:figure1_d}
    \end{subfigure}
    \caption{
    PIDLoc performs localization by incorporating local, global, and fine-grained contexts. In (a)-(c), the red, green, and blue circular sectors represent the current, ground-truth, and predicted pose, respectively. The blue and black arrows represent the position adjustment of the given pose during the single and final iteration, respectively. The yellow region represents the range of poses addressed by each branch. (a) Similar to existing methods, the P branch relies solely on the given pose, often converging to a local optimum. (b) The I branch incorporates global context from diverse poses, enabling robust pose estimation even under large initial pose errors. (c) The D branch leverages gradients of feature differences to perform fine-grained pose adjustments.   
    }
    \label{fig:figure1}
\end{figure}

Cross-view geo-localization has been proposed to address this challenge by leveraging satellite-view images to estimate the global pose of a vehicle. A common approach in this field is cross-view image retrieval~\cite{hu2018cvm, liu2019lending, shi2020optimal, toker2021coming, zhu2021vigor, xia2022visual, xia2023convolutional, lentsch2023slicematch, sarlin2023orienternet}, which matches ground-view images to geo-tagged satellite-view images. However, the localization accuracy is constrained by the partitioning resolution of satellite maps~\cite{shi2022beyond}, often leading to lower precision compared with commercial GNSS~\cite{van2015world}.

Recently, cross-view pose optimization methods~\cite{shi2022beyond, song2024learning, wang2024fine, shi2023boosting, wang2023satellite, wang2023view, wang2024view, shi2025weakly} have been proposed to improve localization accuracy. Given an initial pose roughly estimated from cross-view image retrieval, these methods directly estimate the global pose (position and orientation) of the ground-view image relative to the satellite-view image. These direct pose optimizations enable precise localization beyond the limitations of partitioning resolution~\cite{shi2022beyond}. 

However, existing pose optimization methods heavily rely on cross-view features at a \tcolor{given} pose between satellite-view and ground-view images~\cite{shi2022beyond, wang2023satellite, wang2023view, wang2024view, shi2025weakly}, neglecting the global and fine-grained contexts, similar to the P branch in Fig.~\ref{fig:figure1_p}. This reliance degrades accuracy under large initial pose errors, particularly in environments with repetitive patterns (e.g., recurring buildings or trees), which often leads to convergence to a local optimum. Moreover, these methods fail to capture fine-grained details, such as subtle differences in structural edges or object boundaries, which are critical for precise alignment. Consequently, these methods struggle to achieve accurate feature alignment under large initial pose errors, such as a $40\text{m} \times 40\text{m}$ area.

To overcome these challenges, we propose PIDLoc, a novel approach for cross-view pose optimization that leverages global context and fine-grained local adjustments. As shown in Fig.~\ref{fig:figure1}, the PIDLoc is inspired by the principle of the PID controller~\cite{franklin2002feedback}, a widely used feedback control system in control theory. The PID controller combines proportional (P), integral (I), and derivative (D) components to ensure accurate convergence to a target value. Based on this principle, the PIDLoc interprets cross-view feature relationships through dedicated PID branches, designed as parallel modules within a neural network. Specifically. the P branch adjusts feature differences at the \tcolor{given pose}, the I branch incorporates global context to avoid local optima in repetitive patterns and enhance robustness to large initial pose errors, and the D branch incorporates gradients of feature differences for precise pose refinement.


Integrated with the PID branches, the spatially aware pose estimator (\dnnpose{}) ensures consistent pose estimation by capturing spatial relationships within the PID-branch features. In contrast, existing methods~\cite{shi2022beyond, wang2023satellite, wang2023view, wang2024view} independently estimate each feature's pose, often converging to different local optima and resulting in inconsistent pose estimations when averaged. Our approach addresses these issues by modeling local spatial relationships within the PID-branch features using channel-shared MLPs, effectively encoding a spatial structure for consistent pose estimation. By combining PID branches with the \dnnpose{}, PIDLoc achieves accurate and consistent pose estimation even under large initial pose errors, achieving state-of-the-art results on the cross-view KITTI~\cite{geiger2013vision, shi2022beyond} and Ford Multi-AV Seasonal (FMAVS)~\cite{agarwal2020ford, shi2022beyond, wang2023satellite} datasets.

Our contributions can be summarized as follows:
\begin{itemize}
  \item The PID branches leverage the principle of the PID controller to achieve robust cross-view pose optimization under large initial pose errors.
  \item The proposed \dnnpose{} ensures consistent pose estimation by incorporating the spatial structure of the PID-branch features.
  \item The PIDLoc combines the PID branches and \dnnpose{}, achieving SOTA performances with mean position errors of $4.96\text{m}$ and $4.81\text{m}$ on the cross-view KITTI and FMAVS datasets, respectively.
\end{itemize}

%% file: sec/2_formatting.tex
\section{Related works}
\label{sec:formatting}

\subsection{Cross-view feature matching}
Early cross-view geo-localization methods approached localization as an image retrieval task, achieving broad-scale coverage~\cite{hu2018cvm, liu2019lending, shi2020optimal, toker2021coming, zhu2021vigor, deuser2023sample4geo}. However, these methods often yield lower precision than GNSS~\cite{van2015world} because they assume the estimated position is centered within the image. To enhance localization precision, several works have employed small patch-wise feature matching between ground and satellite views. Xia~\etal~\cite{xia2022visual, xia2023convolutional} employed patch attention to generate dense spatial representations for precise localization. SliceMatch~\cite{lentsch2023slicematch} divides the ground and satellite views into small patches and computes patch-wise similarities with cross-view attention. However, excessively small patches fail to capture the global context, degrading overall performance~\cite{lentsch2023slicematch}.

\subsection{Cross-view pose optimization}
\paragraph{Dense feature-based methods}
Cross-view pose optimization~\cite{shi2022beyond, wang2023satellite, shi2023boosting, wang2023view, wang2024view, song2024learning} directly estimates the camera pose relative to a satellite-view image, avoiding issues with the partitioning resolution of the satellite-view image. Research in this field has primarily focused on mitigating domain gaps arising from perspective, scale, and appearance differences between ground and satellite views. HighlyAccurate~\cite{shi2022beyond} reduces the domain gap using ground homography, but depth ambiguity often leads to misalignments. Subsequent methods~\cite{song2024learning, shi2023boosting} have sought to improve alignment by addressing the limitations of ground homography. Song~\etal~\cite{song2024learning} refined the bird’s-eye view (BEV) feature maps using convolutional neural networks. Boosting~\cite{shi2023boosting} refines BEV feature maps with transformers to capture depth variations in the scene. Nonetheless, they still struggle with feature alignment in complex and non-planar scenes due to persistent depth ambiguity. 

\paragraph{Sparse feature-based methods}
Recent cross-view pose optimization have focused on extracting sparse features based on reliable depth to improve cross-view feature alignments. CSLA~\cite{fervers2022continuous} leverages LiDAR-derived depth for feature alignments but is limited to estimate the position. SIBCL~\cite{wang2023satellite} addresses this by refining both the position and orientation with a Levenberg-Marquardt (LM) optimizer~\cite{more2006levenberg}. The optimizer minimizes residuals between cross-view features at the \tcolor{given} pose to estimate the position and orientation. RGB-based methods like PureACL and VFA~\cite{wang2023view, wang2024view} fine-tune models pre-trained on SIBCL~\cite{wang2023satellite} to extract sparse features that are important for understanding 3D structures. PureACL~\cite{wang2023view} extracts on-ground keypoints for cross-view feature alignments but lacks off-ground keypoints, limiting its effectiveness in complex scenes. VFA~\cite{wang2024view} addresses this by introducing top-down feature aggregation to incorporate off-ground features.

Despite these advancements, cross-view pose optimization methods~\cite{shi2022beyond, wang2023view, wang2024view, shi2023boosting, song2024learning} heavily rely on cross-view features at the \tcolor{given} pose, resulting in performance degradation with increasing initial pose error. To address these issues, we propose a novel PID controller-inspired network that integrates global and fine-grained contexts to ensure accurate localization even under large initial pose errors.

%% file: sec/3_finalcopy.tex
\newcommand{\Dtrain}{(\mathbf{x},y)\sim\mathcal{D}_\text{train}}
\newcommand{\p}[2]{\boldsymbol{p}_{#1}^{#2}}
\newcommand{\ControlFeature}[2]{w_{#1}(\Pose{#2})}
\newcommand{\FeatureError}[1]{e(\Pose{#1})}
\newcommand{\SmallPose}[1]{\mathbf{p}_{#1}}
\newcommand{\Pose}[1]{\mathbf{P}_{#1}}
\newcommand{\SatCoord}{\begin{pmatrix} u^s & v^s \end{pmatrix}^{\intercal}}
\newcommand{\GrdCoord}{\begin{pmatrix} u^g & v^g \end{pmatrix}^{\intercal}}
\newcommand{\SatWCoord}{\begin{pmatrix} x^s & y^s \end{pmatrix}^{\intercal}}
\newcommand{\DoF}{(x,y,\theta)}

\newcommand{\Iset}[1]{\mathcal{I}_{#1}}
\newcommand{\Fg}{\bold{F}_g}
\newcommand{\Fs}{\bold{F}_s}
\newcommand{\trans}{\intercal}
\newcommand{\uvw}{\begin{bmatrix}u & v & 1\end{bmatrix}}
\newcommand{\uv}{\begin{bmatrix}u & v\end{bmatrix}}
\newcommand{\uvT}{\begin{bmatrix}u \\ v\end{bmatrix}}
\newcommand{\xyz}{\begin{bmatrix}x&y&z\end{bmatrix}}
\newcommand{\xyzT}{\begin{bmatrix}x\\y\\z\end{bmatrix}}
\newcommand{\T}{\bold{T}}
\newcommand{\K}[1]{\bold{K}_{#1}}
\newcommand{\pcd}{\mathcal{X}_g}

\begin{figure*}[ht]
    \centering
    \includegraphics[width=\textwidth]{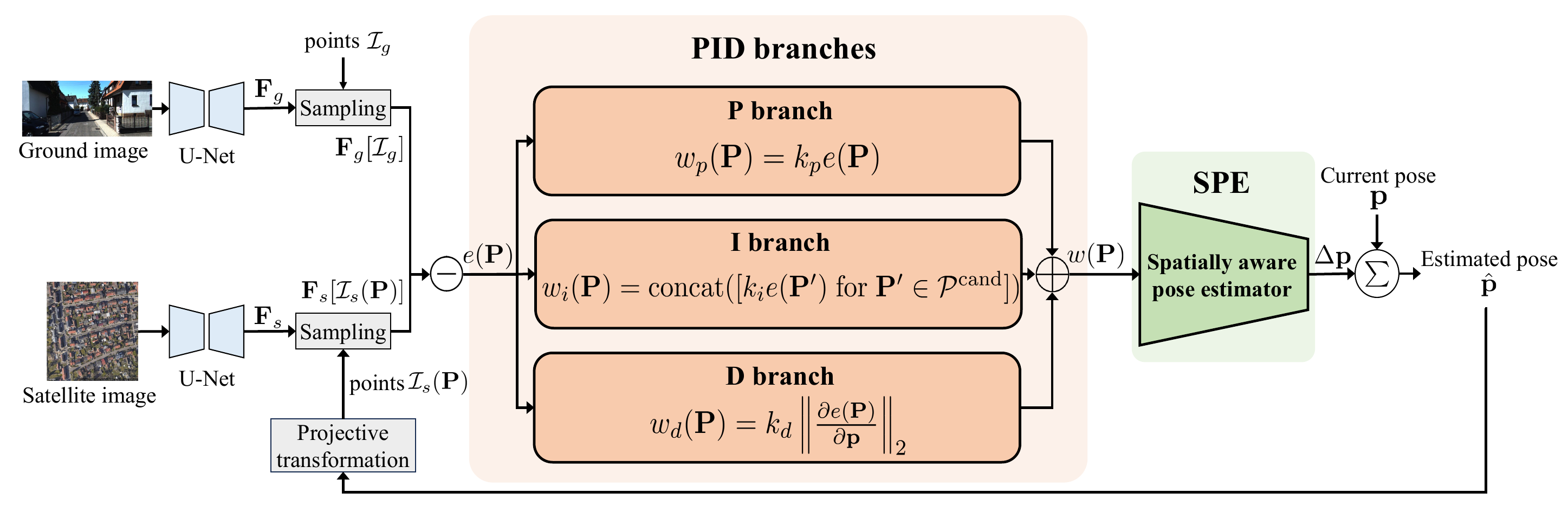} 
    \caption{An overview of the PIDLoc. The PIDLoc iteratively updates the pose based on the cross-view features. The $\oplus$ and $\ominus$ denote concatenation and subtraction, respectively. The proposed method generates the PID-branch features $\ControlFeature{}{}$ from cross-view feature differences $\FeatureError{}=\Fs[\Iset{s}(\Pose{})]-\Fg[\Iset{g}]$. The PID-branch features $\ControlFeature{}{}$ guide the pose estimator \dnnpose{} to converge accurately toward the ground-truth pose even under large initial pose errors.}
    \label{fig:main_figure}
\end{figure*}

\section{The proposed method}
\paragraph{Task settings}
Cross-view pose optimization aims to estimate the 3-DoF vehicle pose $\SmallPose{}=(x,y,\theta)$ relative to the satellite-view image, given an initial coarse pose $\SmallPose{\text{init}}$. The transformation matrix $\Pose{}$, parameterized by $\SmallPose{\text{}}$, maps ground-view camera coordinates to the satellite-view camera coordinates, capturing the vehicle's position $(x,y)$ and orientation $\theta$. Position $(x,y)$ specifies the vehicle's longitudinal and lateral positions, while orientation $\theta$ specifies the azimuth. Consistent with~\cite{shi2022beyond}, we assume that the roll and pitch of the ground-view camera are negligible.

\paragraph{Overview}
This paper proposes PIDLoc, a novel approach to cross-view pose optimization that is robust to large initial pose errors. Existing pose optimization methods primarily emphasize feature extraction to compare ground-view and satellite-view images, \tcolor{In contrast, our approach utilizes LiDAR as~\cite{wang2023satellite} to focus on improving the pose optimization process itself.} As illustrated in Fig.~\ref{fig:main_figure}, our method introduces the PID branches for local, global, and fine-grained contexts; and spatially aware pose estimator (\dnnpose{}) for consistent pose estimation.

\subsection{Cross-view visual feature extraction}   
Ground and satellite-view feature maps denoted as $\Fg$ and $\Fs$, are extracted from ground and satellite-view images using a shared-weight U-Net architecture, respectively. LiDAR points are then projected onto these feature maps to sample corresponding cross-view features. The coordinate set $\Iset{g}$ is computed by projecting the LiDAR points onto the ground-view image:
\begin{equation}
\Iset{g} = \{ (u,v) 
| \uvw^{\trans} = \K{g}\bold{x},\forall \bold{x}\in\pcd
\},
\end{equation}
\noindent where $u$ and $v$ are the image coordinates, $\K{g}$ is the intrinsic matrix of the ground-view camera, and $\pcd$ is the point cloud set of a LiDAR transformed into the ground-view camera coordinate system.
Similarly, the coordinate set $\Iset{s}$ is obtained by projecting the LiDAR points onto the satellite-view image using the transformation matrix $\Pose{}$:
\begin{equation}
    \Iset{s}(\Pose{})=\{(u,v)
    | \uvw^{\trans} 
    =\K{s}\Pose{}\bold{x}, \forall \bold{x}\in\pcd
    \},
\end{equation}
\noindent where $\K{s}$ is the intrinsic matrix of the satellite-view camera. 
Visual features are then sampled from ground and satellite-view image coordinate sets $\Iset{g}$ and $\Iset{s}$. These sampled visual features are used to construct the cross-view features, represented as $\Fg[\Iset{g}]$ and $\Fs[\Iset{s}(\Pose{})]$.

\subsection{PIDLoc}
\subsubsection{A motivation for PID branches}
\tcolor{Existing cross-view pose optimization methods~\cite{shi2022beyond, wang2023view, wang2024view, shi2023boosting, song2024learning} iteratively update their outputs based on the given error, similar to the proportional (P) controller in a feedback control system~\cite{franklin2002feedback}. However, both methods rely only on the given error, neglecting the global and fine-grained contexts. To address the localization problem, we draw inspiration from the PID controller, which incorporate diverse contexts to overcome the limitations of the P controller. The PID controller combines proportional (P), integral (I), and derivative (D) components to generate a control signal $c(t)$ from the error $e(t)$ as follows:
\begin{equation}
c(t)=g_p e(t)+g_i \int e(t) dt +g_d \frac{d e(t)}{dt},
\end{equation}
\noindent where $g_p$, $g_i$, and $g_d$ are the P, I, and D gains, respectively. The control signal $c(t)$ is weighted by three components: the current error $e(t)$, the accumulated past error $\int e(t) dt$, and the rate of change in error $\frac{de(t)}{dt}$, minimizing the target error effectively.}

\subsubsection{PID branches}
\tcolor{Similarly, our proposed PID branches generate features $\ControlFeature{}{}$ from cross-view feature difference $\FeatureError{}$, as shown in Fig.~\ref{fig:main_figure}. The feature difference $\FeatureError{}$ is defined as follows:
\begin{equation}
\FeatureError{}=\Fs[\Iset{s}(\Pose{})]-\Fg[\Iset{g}].
\end{equation}
The PID-branch features $\ControlFeature{}{}$ guide the DNN-based pose estimator to converge accurately by utilizing the feature difference $\FeatureError{}$. The features $\ControlFeature{}{}$ are composed of three components for diverse contexts as follows:
\begin{equation}
\ControlFeature{}{}=\ControlFeature{p}{}\oplus \ControlFeature{i}{} \oplus \ControlFeature{d}{},
\end{equation}
where $\ControlFeature{p}{}$, $\ControlFeature{i}{}$, and $\ControlFeature{d}{}$ are the proportional (P), integral (I), and derivative (D) features, respectively; and $\oplus$ denotes the concatenation. While the PID controller directly adjusts $e(t)$ through $c(t)$, the PIDLoc guides the DNNs to estimate the $\Delta\SmallPose{}$ through $\ControlFeature{}{}$. The specific roles of each branch are detailed below.}

\paragraph{P branch}
The proportional (P) controller generates a control signal $g_p e(t)$ proportional to the current error $e(t)$, enabling rapid error correction. Similarly, the proposed P branch generates the P branch feature $\ControlFeature{p}{}$ in proportion to the feature difference between the ground and satellite views:
\begin{equation}
    \ControlFeature{p}{}=k_p \FeatureError{},
\end{equation}
\noindent where $k_p$ is the learnable proportional coefficient. The P-branch feature guides the pose estimator to adjust the pose, similar to existing methods. However, the P branch relies solely on the feature at the \tcolor{given} pose, restricting its ability to utilize a wider field of view (FoV). This limitation makes the pose estimator prone to local minima and sensitive to the initial pose error. Additionally, the P branch does not incorporate feature difference gradients relative to the pose changes, limiting the precision of the pose estimator.

\paragraph{I branch}
The integral (I) controller generates a control signal $g_i \int e(t) dt$ by accumulating past errors over time, leveraging a broader temporal context to improve accuracy. \tcolor{Based on this concept, the I branch generates the feature $\ControlFeature{i}{}$ by concatenating feature differences of pose candidates $\mathcal{P}^\text{cand}$ along the channel dimension:
\begin{equation}
    \ControlFeature{i}{}=\text{concat}([k_i e(\mathbf{P}')\; \mathrm{for}\; \Pose{}'\in\mathcal{P}^\text{cand}]),
\end{equation}
\noindent where $k_i$ is the learnable integral coefficient. The channel encodes the relative spatial relationships between the given pose and $\mathcal{P}^\text{cand}$. Thus, the I branch enables the pose estimator to update the given pose along the direction corresponding to the channel with the smallest feature error.} The pose candidates $\mathcal{P}^\text{cand}$ are obtained by a grid search over the 3-DoF pose space around the \tcolor{given pose $\SmallPose{} = (x, y, \theta)$.
\begin{equation}
\begin{aligned}
    \mathcal{P}^\text{cand} = 
    \{
    &(x+i s_x, y+j s_y, \theta+k s_\theta) 
    \mid
    s_x|i|\leq r_x,  \\& s_y|j|\leq r_y,
    s_\theta|k|\leq r_k, \forall i,j,k\in\mathbb{Z}\},
\end{aligned}
\end{equation}
\noindent where $r_x, r_y$, and $r_z$ are the search radius along each axis; $s_x, s_y$, and $s_\theta$ represent the step size for the grid search. The $\mathcal{P}^\text{cand}$ enables the pose estimator to compare feature differences within a global spatial context, effectively addressing the limited FoV of the P branch.} Additionally, gradient clipping is applied to the network parameters to prevent excessive feature magnitudes from concatenation, analogous to the anti-windup mechanism in an I controller.

Repetitive patterns along the longitudinal directions, such as recurring buildings or trees, pose significant challenges for cross-view pose optimization. Existing methods rely solely on the \tcolor{given} pose, struggling to distinguish repetitive patterns and often converging to a local optimum. \tcolor{In contrast, the I branch utilizes multiple pose candidates to incorporate global context, reducing sensitivity to repetitive patterns and improving robustness to initial pose errors. Consequently, it achieves a more accurate pose estimation compared with using the P branch alone.}

\paragraph{D branch}
The derivative (D) controller generates a control signal $g_d \frac{de(t)}{dt}$ based on the rate of error change, allowing precise adjustment. Similarly, the proposed D branch generates the feature $\ControlFeature{d}{}$ by considering feature difference gradients relative to the pose changes:
\newcommand{\norm}[1]{\left\lVert#1\right\rVert}
\begin{equation}
    \ControlFeature{d}{}=k_d 
    \norm{\frac{\partial \FeatureError{}}{\partial \SmallPose{}}}_2,
\end{equation}
\noindent where $k_d$ is the learnable derivative coefficient. An $\ell_2\text{-norm}$ is applied to enhance robustness against high-frequency noise. \tcolor{The D branch captures the sensitivity of feature differences $\FeatureError{}$ to small spatial variations of a given pose $\SmallPose{}=(x,y,\theta)$, enabling the pose estimator to adjust the pose more precisely. The feature difference gradients are computed via multi-variable differentiation using gradient descent as follows:}
\begin{equation}
    \frac{\partial e(\Pose{})}{\partial \SmallPose{}} = \frac{\partial (\Fs [\SatCoord])}{\partial \SatCoord}  
        \frac{\partial \SatCoord}{\partial \SatWCoord}  \frac{\partial \SatWCoord}{\partial \SmallPose{}},
\end{equation}
where $\SmallPose{}$ is the current 3-DoF pose $(x, y, \theta)$; $\SatCoord$ and $\SatWCoord$ are the image and camera coordinates in the satellite-view, respectively. The feature difference gradients from each direction are summed to produce the aggregated D-branch feature $\ControlFeature{d}{}$.

Existing DNN-based methods~\cite{lentsch2023slicematch, song2024learning, shi2023boosting} mainly rely on the cross-view features at a given pose, limiting their ability to capture subtle differences in structural edges or object boundaries. By contrast, our D branch leverages feature difference gradients across $(x, y, \theta)$ to capture subtle feature variations for the fine-grained alignment, enabling precise pose estimation.

\subsubsection{Spatially aware pose estimator with PID-branch features}
\begin{figure}[!t]
    \centering
    \includegraphics[width=1.0\columnwidth]{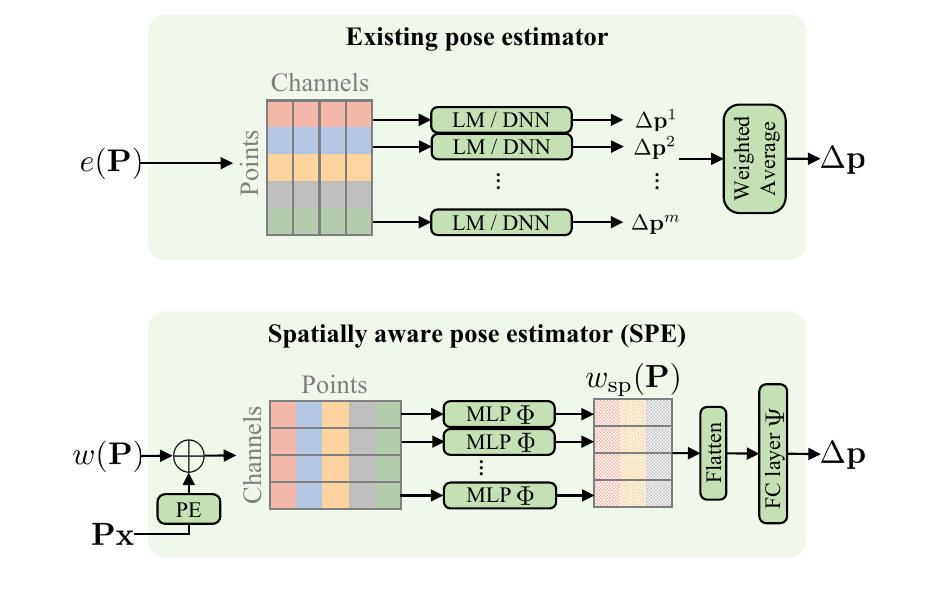} 
    \caption{The comparison of existing pose estimators and proposed SPE. $\Delta \textbf{p}^m$ is the pose difference of the point $m$, $\text{PE}$ is a positional embedding, $\ControlFeature{}{}$ are PID-branch features, $\Pose{} \textbf{x}$ are 3D satellite-view coordinates, and MLP $\Phi$ are channel-shared MLPs.}
    \label{fig:pose_estimator}
\end{figure}

The proposed spatially aware pose estimator (SPE) incorporates spatial relationships among the PID-branch features to achieve consistent pose estimation. As shown in Fig.~\ref{fig:pose_estimator}, recent methods~\cite{shi2022beyond, wang2023satellite, wang2023view, wang2024view} independently estimate the pose for each feature corresponding to individual coordinates and then average the results. Estimating the pose independently can overlook spatial correlations among the features, such as the structural relationships within the building walls or between building walls and road edges. 
In contrast, \dnnpose{} explicitly models these spatial dependencies, resulting in more accurate and consistent pose estimation.

\newcommand{\RGB}{C}
\newcommand{\LiDAR}{L}
\newcommand{\Lp}{Lp}

\begin{table*}[ht!]  
    \begin{centering}
    \begin{adjustbox}{center, max width=0.96\textwidth}
    \begin{threeparttable}
        \footnotesize{
        \begin{tabular}
            {@{}r  *{1}{c}@{} *{3}{c}@{} | *{4}{c}@{} | *{4}{c}@{} | *{4}{c}@{} | *{1}{c}@{}}
            \\ 
            \toprule 
                \multicolumn{1}{c|@{}}{\multirow{3}{*}{Method}}
                & \multicolumn{1}{c|@{}}{\multirow{3}{*}{Area}}
                & \multicolumn{1}{c|@{}}{\multirow{3}{*}{Modalities}}
                & \multicolumn{6}{c@{}}{\raisebox{0.5ex}{Position}}
                & \multicolumn{4}{c@{}}{\raisebox{0.5ex}{Orientation}}
            \\ \noalign{\vskip -0.85mm}
            \cmidrule(lr){4-9} \cmidrule(lr){10-13}  \noalign{\vskip -0.85mm}
                \multicolumn{1}{c|@{}}{}
                & \multicolumn{1}{c|@{}}{}
                & \multicolumn{1}{c|@{}}{}
                & \multicolumn{2}{c|@{}}{Error (m) ↓} 
                & \multicolumn{2}{c|@{}}{Lateral recall (\%) ↑}
                & \multicolumn{2}{c|@{}}{Long. recall (\%) ↑}
                & \multicolumn{2}{c|@{}}{Error ($^{\circ}$) ↓} 
                & \multicolumn{2}{c@{}}{Recall (\%) ↑}
            \\
                \multicolumn{1}{c|@{}}{}
                & \multicolumn{1}{c|@{}}{}
                & \multicolumn{1}{c|@{}}{}
                & \mc{Mean} & \multicolumn{1}{c|@{}}{Med} 
                & \mc{R@1m} & \multicolumn{1}{c|@{}}{R@5m}
                & \mc{R@1m} & \multicolumn{1}{c|@{}}{R@5m}
                & \mc{Mean} & \multicolumn{1}{c|@{}}{Med} 
                & \mc{R@1$^{\circ}$} & \mc{R@5$^{\circ}$}
            \\ \noalign{\vskip 0.5mm}
            \hline \noalign{\vskip 0.5mm}
                \multicolumn{1}{r|@{}}{DSM~\cite{shi2020looking}}
                & \multicolumn{1}{c|@{}}{Same}
                & \multicolumn{1}{c|@{}}{\RGB}
                & \mc{-} & \multicolumn{1}{c|@{}}{-} 
                & \mc{10.12} & \multicolumn{1}{c|@{}}{48.24}
                & \mc{4.08} & \multicolumn{1}{c|@{}}{20.14}
                & \mc{-} & \multicolumn{1}{c|@{}}{-} 
                & \mc{3.58} & \mc{24.4}
            \\ 
                \multicolumn{1}{r|@{}}{HighlyAccurate~\cite{shi2022beyond}}
                & \multicolumn{1}{c|@{}}{Same}
                & \multicolumn{1}{c|@{}}{\RGB}
                & \mc{12.08} & \multicolumn{1}{c|@{}}{11.42} 
                & \mc{35.54} & \multicolumn{1}{c|@{}}{80.36}
                & \mc{5.22} & \multicolumn{1}{c|@{}}{26.31}
                & \mc{3.72} & \multicolumn{1}{c|@{}}{2.83} 
                & \mc{19.64} & \mc{71.72}
            \\ 
            \multicolumn{1}{r|@{}}{SliceMatch~\cite{lentsch2023slicematch}}
                & \multicolumn{1}{c|@{}}{Same}
                & \multicolumn{1}{c|@{}}{\RGB}
                & \mc{7.96} & \multicolumn{1}{c|@{}}{4.39} 
                & \mc{49.09} & \multicolumn{1}{c|@{}}{98.52}
                & \mc{15.19} & \multicolumn{1}{c|@{}}{57.35}
                & \mc{4.12} & \multicolumn{1}{c|@{}}{3.65} 
                & \mc{13.41} & \mc{64.17}
            \\ 
                 \multicolumn{1}{r|@{}}{PureACL$\dagger$~\cite{wang2023view}}
                & \multicolumn{1}{c|@{}}{Same}
                & \multicolumn{1}{c|@{}}{\RGB\ (\Lp)}
                & \mc{11.57} & \multicolumn{1}{c|@{}}{11.33}  
                & \mc{29.87} & \multicolumn{1}{c|@{}}{79.80}  
                & \mc{4.72} & \multicolumn{1}{c|@{}}{25.50}   
                & \mc{5.03} & \multicolumn{1}{c|@{}}{5.05} 
                & \mc{9.89} & \mc{49.40} 
            \\ 
                 \multicolumn{1}{r|@{}}{VFA$\dagger$~\cite{wang2024view}}
                & \multicolumn{1}{c|@{}}{Same}
                & \multicolumn{1}{c|@{}}{\RGB\ (\Lp)}
                & \mc{-} & \multicolumn{1}{c|@{}}{-}  
                & \mc{-} & \multicolumn{1}{c|@{}}{-}  
                & \mc{-} & \multicolumn{1}{c|@{}}{-}   
                & \mc{-} & \multicolumn{1}{c|@{}}{-} 
                & \mc{-} & \mc{-} 
            \\ 
            \multicolumn{1}{r|@{}}{SIBCL$\dagger$~\cite{wang2023satellite}}
                & \multicolumn{1}{c|@{}}{Same}
                & \multicolumn{1}{c|@{}}{\RGB\ +\ \LiDAR}
                & \mc{9.14} & \multicolumn{1}{c|@{}}{7.43} 
                & \mc{46.59} & \multicolumn{1}{c|@{}}{87.44}
                & \mc{11.07} & \multicolumn{1}{c|@{}}{39.33}
                & \mc{3.15} & \multicolumn{1}{c|@{}}{2.13} 
                & \mc{27.09} & \mc{78.53}
            \\  
            \multicolumn{1}{r|@{}}{Boosting~\cite{shi2023boosting}}
                & \multicolumn{1}{c|@{}}{Same}
                & \multicolumn{1}{c|@{}}{\RGB}
                & \mc{10.01} & \multicolumn{1}{c|@{}}{5.19} 
                & \mc{76.44} & \multicolumn{1}{c|@{}}{98.89}
                & \mc{23.54} & \multicolumn{1}{c|@{}}{62.18}
                & \mc{0.55} & \multicolumn{1}{c|@{}}{0.42} 
                & \mc{99.10} & \mc{\textbf{100.00}}
            \\  
            \multicolumn{1}{r|@{}}{CCVPE~\cite{xia2023convolutional}}
                & \multicolumn{1}{c|@{}}{Same}
                & \multicolumn{1}{c|@{}}{\RGB}
                & \mc{1.22} & \multicolumn{1}{c|@{}}{0.62} 
                & \mc{97.35} & \multicolumn{1}{c|@{}}{99.71}
                & \mc{77.13} & \multicolumn{1}{c|@{}}{97.16}
                & \mc{0.67} & \multicolumn{1}{c|@{}}{0.54} 
                & \mc{77.39} & \mc{99.95}
            \\  
            \multicolumn{1}{r|@{}}{Song~\etal~\cite{song2024learning}}
                & \multicolumn{1}{c|@{}}{Same}
                & \multicolumn{1}{c|@{}}{\RGB}
                & \mc{1.48} & \multicolumn{1}{c|@{}}{0.47} 
                & \mc{95.47} & \multicolumn{1}{c|@{}}{\textbf{99.79}}
                & \mc{87.89} & \multicolumn{1}{c|@{}}{94.78}
                & \mc{0.49} & \multicolumn{1}{c|@{}}{0.30} 
                & \mc{89.40} & \mc{99.31}
            \\ 
            \noalign{\vskip 0.7mm}
                \rowcolor[gray]{0.9}
                \multicolumn{1}{r|@{}}{\textbf{PIDLoc (Ours)}}
                & \multicolumn{1}{c|@{}}{Same}
                & \multicolumn{1}{c|@{}}{\RGB\ +\ \LiDAR}
                & \mc{\textbf{1.15}} & \multicolumn{1}{c|@{}}{\textbf{0.36}} 
                & \mc{\textbf{97.37}} & \multicolumn{1}{c|@{}}{99.36}
                & \mc{\textbf{94.22}} & \multicolumn{1}{c|@{}}{\textbf{97.61}}
                & \mc{\textbf{0.11}} & \multicolumn{1}{c|@{}}{\textbf{0.09}} 
                & \mc{\textbf{99.97}} & \mc{\textbf{100.00}}
            \\ 
            \midrule
            \hline \noalign{\vskip 0.7mm}
                \multicolumn{1}{r|@{}}{DSM~\cite{shi2020looking}}
                & \multicolumn{1}{c|@{}}{Cross}
                & \multicolumn{1}{c|@{}}{\RGB}
                & \mc{-} & \multicolumn{1}{c|@{}}{-} 
                & \mc{10.77} & \multicolumn{1}{c|@{}}{48.24}
                & \mc{3.87} & \multicolumn{1}{c|@{}}{19.50}
                & \mc{-} & \multicolumn{1}{c|@{}}{-} 
                & \mc{3.53} & \mc{23.95}
            \\ 
                \multicolumn{1}{r|@{}}{HighlyAccurate~\cite{shi2022beyond}}
                & \multicolumn{1}{c|@{}}{Cross}
                & \multicolumn{1}{c|@{}}{\RGB}
                & \mc{12.58} & \multicolumn{1}{c|@{}}{12.11} 
                & \mc{27.82} & \multicolumn{1}{c|@{}}{72.89}
                & \mc{5.75} & \multicolumn{1}{c|@{}}{26.48}
                & \mc{3.95} & \multicolumn{1}{c|@{}}{3.03} 
                & \mc{18.42} & \mc{71.00}
            \\ 
            \multicolumn{1}{r|@{}}{SliceMatch~\cite{lentsch2023slicematch}}
                & \multicolumn{1}{c|@{}}{Cross}
                & \multicolumn{1}{c|@{}}{\RGB}
                & \mc{13.50} & \multicolumn{1}{c|@{}}{9.77} 
                & \mc{32.43} & \multicolumn{1}{c|@{}}{86.44}
                & \mc{8.30} & \multicolumn{1}{c|@{}}{35.57}
                & \mc{4.20} & \multicolumn{1}{c|@{}}{6.61} 
                & \mc{46.82} & \mc{46.82}
            \\ 
                \multicolumn{1}{r|@{}}{PureACL$\dagger$~\cite{wang2023view}}
                & \multicolumn{1}{c|@{}}{Cross}
                & \multicolumn{1}{c|@{}}{\RGB\ (\Lp)}
                & \mc{11.71} & \multicolumn{1}{c|@{}}{11.57}  
                & \mc{28.55} & \multicolumn{1}{c|@{}}{78.22}  
                & \mc{5.31} & \multicolumn{1}{c|@{}}{25.76}   
                & \mc{5.02} & \multicolumn{1}{c|@{}}{5.01} 
                & \mc{9.56} & \mc{49.92} 
             \\ 
                 \multicolumn{1}{r|@{}}{VFA$\dagger$~\cite{wang2024view}}
                & \multicolumn{1}{c|@{}}{Cross}
                & \multicolumn{1}{c|@{}}{\RGB\ (\Lp)}
                & \mc{-} & \multicolumn{1}{c|@{}}{-}  
                & \mc{-} & \multicolumn{1}{c|@{}}{-}  
                & \mc{-} & \multicolumn{1}{c|@{}}{-}   
                & \mc{-} & \multicolumn{1}{c|@{}}{-} 
                & \mc{-} & \mc{-} 
             \\ 
                \multicolumn{1}{r|@{}}{SIBCL$\dagger$~\cite{wang2023satellite}}
                & \multicolumn{1}{c|@{}}{Cross}
                & \multicolumn{1}{c|@{}}{\RGB\ +\ \LiDAR}
                & \mc{9.73} & \multicolumn{1}{c|@{}}{8.24} 
                & \mc{37.52} & \multicolumn{1}{c|@{}}{78.06}
                & \mc{9.97} & \multicolumn{1}{c|@{}}{36.66}
                & \mc{3.66} & \multicolumn{1}{c|@{}}{2.73} 
                & \mc{20.27} & \mc{74.13}
            \\  
            \multicolumn{1}{r|@{}}{Boosting~\cite{shi2023boosting}}
                & \multicolumn{1}{c|@{}}{Cross}
                & \multicolumn{1}{c|@{}}{\RGB}
                & \mc{13.01} & \multicolumn{1}{c|@{}}{9.06} 
                & \mc{57.72} & \multicolumn{1}{c|@{}}{91.16}
                & \mc{14.15} & \multicolumn{1}{c|@{}}{45.00}
                & \mc{0.56} & \multicolumn{1}{c|@{}}{0.43} 
                & \mc{98.98} & \mc{\textbf{100.00}}
            \\ 
            {CCVPE~\cite{xia2023convolutional}}
                & \multicolumn{1}{c|@{}}{Cross}
                & \multicolumn{1}{c|@{}}{\RGB}
                & \mc{9.16} & \multicolumn{1}{c|@{}}{3.33} 
                & \mc{44.06} & \multicolumn{1}{c|@{}}{90.23}
                & \mc{23.08} & \multicolumn{1}{c|@{}}{64.31}
                & \mc{1.55} & \multicolumn{1}{c|@{}}{0.84} 
                & \mc{57.72} & \mc{96.19}
            \\  
            \multicolumn{1}{r|@{}}{Song~\etal~\cite{song2024learning}}
                & \multicolumn{1}{c|@{}}{Cross}
                & \multicolumn{1}{c|@{}}{\RGB}
                & \mc{7.97} & \multicolumn{1}{c|@{}}{3.52} 
                & \mc{54.19} & \multicolumn{1}{c|@{}}{91.74}
                & \mc{23.10} & \multicolumn{1}{c|@{}}{61.75}
                & \mc{2.17} & \multicolumn{1}{c|@{}}{1.21} 
                & \mc{43.44} & \mc{89.31}
            \\ 
            \noalign{\vskip 0.7mm}
                \rowcolor[gray]{0.9}
                \multicolumn{1}{r|@{}}{\textbf{PIDLoc (Ours)}}
                & \multicolumn{1}{c|@{}}{Cross}
                & \multicolumn{1}{c|@{}}{\RGB\ +\ \LiDAR}
                & \mc{\textbf{4.96}} & \multicolumn{1}{c|@{}}{\textbf{1.17}} 
                & \mc{\textbf{71.01}} & \multicolumn{1}{c|@{}}{\textbf{94.27}}
                & \mc{\textbf{50.02}} & \multicolumn{1}{c|@{}}{\textbf{75.43}}
                & \mc{\textbf{0.17}} & \multicolumn{1}{c|@{}}{\textbf{0.12}} 
                & \mc{\textbf{99.96}} & \mc{\textbf{100.00}}
            \\ 
            \bottomrule
        \end{tabular}
        }
    \end{threeparttable}
    \end{adjustbox}
    \caption{Comparison on the KITTI dataset under $\pm 20 \text{m}$ position noise and $\pm 10^\circ$ orientation noise. \RGB, 4\RGB, \LiDAR, and \Lp\ denote 1 RGB camera, 4 RGB cameras, LiDAR, and LiDAR-pretrained model, respectively. $\dagger$ indicates reproduced results under settings in~\cite{shi2022beyond}}
    \label{tab:shi}
    \end{centering}
\end{table*} 

The SPE employs channel-shared multi-layer-perceptrons (MLPs) $\Phi$ to generate features $\ControlFeature{\text{sp}}{}$, incorporating the spatial dependencies of PID-branch features:
\begin{equation}
    \ControlFeature{\text{sp}}{} = \Phi(\ControlFeature{}{} \oplus \textbf{PE}(\Pose{} \textbf{x})),
\end{equation}
where $\textbf{PE}$ represents a 2-layer MLP for \tcolor{positional embedding} and $\Pose{} \textbf{x}$ denote 3D satellite-view coordinates because $\textbf{x}$ represent the transformed LiDAR points in ground-view camera coordinates. $\ControlFeature{}{}$ is concatenated with \tcolor{positional embeddings} $\textbf{PE}(\Pose{} \textbf{x})$, creating a position-dependent representation essential for pose estimation. 

Subsequently, the positionally encoded PID-branch features are fed into channel-shared MLPs $\Phi$, which apply the same parameters across all channels. The channel-shared MLPs efficiently learn the spatial relationships among features, generating a compact representation $\ControlFeature{\text{sp}}{}$ with reduced spatial dimensionality. Through this process, each feature point interacts spatially to form an aggregated representation that captures the underlying spatial structure. 
Leveraging these spatial dependencies, the MLPs $\Phi$ ensures robust and consistent pose estimation.

Finally, $\ControlFeature{\text{sp}}{}$ is flattened across the channel and spatial dimensions, followed by a fully connected layer $\Psi$ to predict the pose difference $\Delta\SmallPose{}\in \mathbb{R}^{3}$ for pose refinement.
\begin{equation}
    \Delta\SmallPose{} = \Psi(\text{Flatten}(\ControlFeature{\text{sp}}{})).
\end{equation}
Because the $\ControlFeature{sp}{}$ is positionally encoded, $\Psi$ can leverage the diverse local features to predict the pose difference, improving the precision of pose estimation. 
The estimated pose $\hat{\SmallPose{}}=(\hat{x}, \hat{y}, \hat{\theta})$, calculated as $\hat{\SmallPose{}}=\SmallPose{} + \Delta{\SmallPose{}}$, is trained with an $\ell_1\text{-loss}$ against the ground-truth pose as follows:
\begin{equation}
    L = \sum_{l}(||\hat{x}^l-x^*||_1+||\hat{y}^l-y^*||_1+||\hat{\theta}^{l}-\theta^*||_1), 
\end{equation}
where $(\hat{x}^l, \hat{y}^l, \hat{\theta}^{l})$ denotes the estimated pose (latitude, longitude, and azimuth) at each feature level $l$ and $(x^*, y^*, \theta^*)$ denotes the ground-truth pose.


%% file: sec/4_experiments.tex
\section{Experiments}
\label{sec:experiments}
\paragraph{Datasets}
We evaluated the proposed method against two well-known autonomous driving datasets: KITTI dataset~\cite{geiger2013vision} and Ford Multi-AV Seasonal dataset (FMAVS)~\cite{agarwal2020ford}, consistent with the experimental settings in HighlyAccurate~\cite{shi2022beyond} and SIBCL~\cite{wang2023satellite}, respectively. \tcolor{Satellite-view images were retrieved from Google Maps using GPS tags with a size of $512 \times 512$ pixels. The ground sampling resolutions are $0.2\text{m}$ per pixel for KITTI and $0.22\text{m}$ per pixel for FMAVS datasets, respectively.} For both datasets, experiments were conducted in two distinct setups~\cite{shi2022beyond, wang2023satellite}: same-area and cross-area. In the same-area setup, samples were drawn from the same regions as the training data, while in the cross-area setup, samples were drawn from different regions. The initial pose error was set to $\pm 20\text{m}$ for position and $\pm 10^\circ$ for orientation unless otherwise specified.

\paragraph{Implementation details}
\tcolor{Consistent with SIBCL~\cite{wang2023satellite}, we adopt a U-Net backbone with a VGG-16 encoder~\cite{simonyan2014very} pre-trained on ImageNet~\cite{deng2009imagenet} and a randomly initialized decoder. We randomly sample 5,000 LiDAR points within the ground-view camera's field of view (FoV). The training was conducted on an NVIDIA RTX A5000 GPU with a batch size of 4, using the Adam optimizer~\cite{KingBa15} with a learning rate of $10^{-4}$ for 30 epochs. The PID branches' coefficients $k_p$, $k_i$, and $k_d$ are initialized to $1$ and optimized with a learning rate of $10^{-4}$. Pose optimization is performed across three levels of feature maps with five iterations for each level. In the I branch, the search radius and step size for pose candidates are set to half of the initial pose noise radius, resulting in two samples per direction. The inference time of the PIDLoc is $370\text{ms}$, outperforming $500\text{ms}$ of the HighlyAccurate~\cite{shi2022beyond}.} 

Furthermore, previous studies~\cite{wang2023satellite, wang2023view} aligned ground-truth poses to the satellite image center, which risks overfitting by biasing predictions towards the center. To ensure a fair comparison, we reproduced the studies without this adjustment. Further details are provided in the supplementary material. 

\paragraph{Evaluation metrics}
We aim to estimate the 3-DoF pose $\SmallPose{}=(x,y,\theta)$, comprised of lateral, longitudinal, and azimuth components. We evaluated the accuracy by measuring the recall rates ($\%$) within $1\text{m}$ and $5\text{m}$ (position) and $1^\circ$ and $5^\circ$ (orientation), along with the mean and median errors in position ($\text{m})$ and orientation ($^\circ)$. 

\subsection{Comparison with the state-of-the-art}
\begin{table*}[ht!]  
    \begin{centering}
    \begin{adjustbox}{center, max width=0.96\textwidth}
    \begin{threeparttable}
        \footnotesize{
        \begin{tabular}
            {@{}r  *{1}{c}@{} *{3}{c}@{} | *{4}{c}@{} | *{4}{c}@{} | *{4}{c}@{} | *{1}{c}@{}}
            \\ 
            \toprule
                \multicolumn{1}{c|@{}}{\multirow{3}{*}{Method}}
                & \multicolumn{1}{c|@{}}{\multirow{3}{*}{Area}}
                & \multicolumn{1}{c|@{}}{\multirow{3}{*}{Modalities}}
                & \multicolumn{6}{c@{}}{\raisebox{0.5ex}{Position}}
                & \multicolumn{4}{c@{}}{\raisebox{0.5ex}{Orientation}}
            \\ \noalign{\vskip -0.85mm}
            \cmidrule(lr){4-9} \cmidrule(lr){10-13}  \noalign{\vskip -0.85mm}
                \multicolumn{1}{c|@{}}{}
                & \multicolumn{1}{c|@{}}{}
                & \multicolumn{1}{c|@{}}{}
                & \multicolumn{2}{c|@{}}{Error (m) ↓} 
                & \multicolumn{2}{c|@{}}{Lateral recall (\%) ↑}
                & \multicolumn{2}{c|@{}}{Long. recall (\%) ↑}
                & \multicolumn{2}{c|@{}}{Error ($^{\circ}$) ↓} 
                & \multicolumn{2}{c@{}}{Recall (\%) ↑}
            \\
                \multicolumn{1}{c|@{}}{}
                & \multicolumn{1}{c|@{}}{}
                & \multicolumn{1}{c|@{}}{}
                & \mc{Mean} & \multicolumn{1}{c|@{}}{Med} 
                & \mc{R@1m} & \multicolumn{1}{c|@{}}{R@5m}
                & \mc{R@1m} & \multicolumn{1}{c|@{}}{R@5m}
                & \mc{Mean} & \multicolumn{1}{c|@{}}{Med} 
                & \mc{R@1$^{\circ}$} & \mc{R@5$^{\circ}$}
            \\ \noalign{\vskip 0.5mm}
            \hline \noalign{\vskip 0.5mm}
                \multicolumn{1}{r|@{}}{HighlyAccurate~\cite{shi2022beyond}}
                & \multicolumn{1}{c|@{}}{Same}
                & \multicolumn{1}{c|@{}}{\RGB}
                & \mc{5.12} & \multicolumn{1}{c|@{}}{4.18} 
                & \mc{20.20} & \multicolumn{1}{c|@{}}{72.96}
                & \mc{25.20} & \multicolumn{1}{c|@{}}{82.49}
                & \mc{5.03} & \multicolumn{1}{c|@{}}{3.91} 
                & \mc{14.48} & \mc{59.60}
            \\ 
            \multicolumn{1}{r|@{}}{PureACL$\dagger$~\cite{wang2023view}}
                & \multicolumn{1}{c|@{}}{Same}
                & \multicolumn{1}{c|@{}}{\RGB\ (\Lp)}
                & \mc{5.45} & \multicolumn{1}{c|@{}}{4.68}  
                & \mc{16.74} & \multicolumn{1}{c|@{}}{70.70}  
                & \mc{20.57} & \multicolumn{1}{c|@{}}{80.61}   
                & \mc{4.88} & \multicolumn{1}{c|@{}}{4.64} 
                & \mc{11.49} & \mc{53.80} 
            \\ 
            \multicolumn{1}{r|@{}}{PureACL$\dagger$~\cite{wang2023view}}
                & \multicolumn{1}{c|@{}}{Same}
                & \multicolumn{1}{c|@{}}{4\RGB\ (\Lp)}
                & \mc{3.43} & \multicolumn{1}{c|@{}}{2.08}  
                & \mc{38.09} & \multicolumn{1}{c|@{}}{85.49}  
                & \mc{48.51} & \multicolumn{1}{c|@{}}{91.17}   
                & \mc{3.16} & \multicolumn{1}{c|@{}}{2.86} 
                & \mc{42.03} & \mc{83.63} 
            \\ 
                 \multicolumn{1}{r|@{}}{VFA$\dagger$~\cite{wang2024view}}
                & \multicolumn{1}{c|@{}}{Same}
                & \multicolumn{1}{c|@{}}{\RGB\ (\Lp)}
                & \mc{-} & \multicolumn{1}{c|@{}}{-}  
                & \mc{-} & \multicolumn{1}{c|@{}}{-}  
                & \mc{-} & \multicolumn{1}{c|@{}}{-}   
                & \mc{-} & \multicolumn{1}{c|@{}}{-} 
                & \mc{-} & \mc{-} 
            \\ 
            \multicolumn{1}{r|@{}}{SIBCL$\dagger$~\cite{wang2023satellite}}
                & \multicolumn{1}{c|@{}}{Same}
                & \multicolumn{1}{c|@{}}{\RGB\ +\ \LiDAR}
                & \mc{4.19} & \multicolumn{1}{c|@{}}{1.08} 
                & \mc{51.40} & \multicolumn{1}{c|@{}}{87.92}
                & \mc{55.47} & \multicolumn{1}{c|@{}}{87.87}
                & \mc{2.96} & \multicolumn{1}{c|@{}}{0.80} 
                & \mc{43.30} & \mc{86.15}
            \\  
            \noalign{\vskip 0.7mm}
                \rowcolor[gray]{0.9}
                \multicolumn{1}{r|@{}}{\textbf{PIDLoc (Ours)}}
                & \multicolumn{1}{c|@{}}{Same}
                & \multicolumn{1}{c|@{}}{\RGB\ +\ \LiDAR}
                & \mc{\textbf{1.42}} & \multicolumn{1}{c|@{}}{\textbf{0.88}} 
                & \mc{\textbf{82.37}} & \multicolumn{1}{c|@{}}{\textbf{97.82}}
                & \mc{\textbf{74.06}} & \multicolumn{1}{c|@{}}{\textbf{96.81}}
                & \mc{\textbf{0.86}} & \multicolumn{1}{c|@{}}{\textbf{0.08}} 
                & \mc{\textbf{65.77}} & \mc{\textbf{90.16}}
            \\ 
            \midrule
                       \hline \noalign{\vskip 0.7mm}
                \multicolumn{1}{r|@{}}{HighlyAccurate~\cite{shi2022beyond}}
                & \multicolumn{1}{c|@{}}{Cross}
                & \multicolumn{1}{c|@{}}{\RGB}
                & \mc{6.58} & \multicolumn{1}{c|@{}}{5.62} 
                & \mc{16.35} & \multicolumn{1}{c|@{}}{74.33}
                & \mc{11.30} & \multicolumn{1}{c|@{}}{65.03}
                & \mc{6.04} & \multicolumn{1}{c|@{}}{4.99} 
                & \mc{11.30} & \mc{50.0}
            \\ 
            \multicolumn{1}{r|@{}}{PureACL$\dagger$~\cite{wang2023view}}
                & \multicolumn{1}{c|@{}}{Cross}
                & \multicolumn{1}{c|@{}}{\RGB\ (\Lp)}
                & \mc{6.50} & \multicolumn{1}{c|@{}}{6.33}  
                & \mc{14.64} & \multicolumn{1}{c|@{}}{64.86}  
                & \mc{13.66} & \multicolumn{1}{c|@{}}{66.49}   
                & \mc{5.35} & \multicolumn{1}{c|@{}}{4.97} 
                & \mc{10.78} & \mc{50.32} 
            \\ 
            \multicolumn{1}{r|@{}}{PureACL$\dagger$~\cite{wang2023view}}
                & \multicolumn{1}{c|@{}}{Cross}
                & \multicolumn{1}{c|@{}}{4\RGB\ (\Lp)}
                & \mc{6.43} & \multicolumn{1}{c|@{}}{4.87}  
                & \mc{18.48} & \multicolumn{1}{c|@{}}{73.70}  
                & \mc{24.33} & \multicolumn{1}{c|@{}}{70.26}   
                & \mc{5.66} & \multicolumn{1}{c|@{}}{4.01} 
                & \mc{14.98} & \mc{58.86} 
            \\ 
                 \multicolumn{1}{r|@{}}{VFA$\dagger$~\cite{wang2024view}}
                & \multicolumn{1}{c|@{}}{Cross}
                & \multicolumn{1}{c|@{}}{\RGB\ (\Lp)}
                & \mc{-} & \multicolumn{1}{c|@{}}{-}  
                & \mc{-} & \multicolumn{1}{c|@{}}{-}  
                & \mc{-} & \multicolumn{1}{c|@{}}{-}   
                & \mc{-} & \multicolumn{1}{c|@{}}{-} 
                & \mc{-} & \mc{-} 
            \\ 
            \multicolumn{1}{r|@{}}{SIBCL$\dagger$~\cite{wang2023satellite}}
                & \multicolumn{1}{c|@{}}{Cross}
                & \multicolumn{1}{c|@{}}{\RGB\ +\ \LiDAR}
                & \mc{15.20} & \multicolumn{1}{c|@{}}{13.69} 
                & \mc{14.24} & \multicolumn{1}{c|@{}}{45.00}
                & \mc{8.43} & \multicolumn{1}{c|@{}}{38.70}
                & \mc{8.96} & \multicolumn{1}{c|@{}}{6.19} 
                & \mc{12.60} & \mc{43.75}
            \\  
            \noalign{\vskip 0.7mm}
                \rowcolor[gray]{0.9}
                \multicolumn{1}{r|@{}}{\textbf{PIDLoc (Ours)}}
                & \multicolumn{1}{c|@{}}{Cross}
                & \multicolumn{1}{c|@{}}{\RGB\ +\ \LiDAR}
                & \mc{\textbf{4.81}} & \multicolumn{1}{c|@{}}{\textbf{2.95}} 
                & \mc{\textbf{41.92}} & \multicolumn{1}{c|@{}}{\textbf{78.69}}
                & \mc{\textbf{29.03}} & \multicolumn{1}{c|@{}}{\textbf{86.06}}
                & \mc{\textbf{2.49}} & \multicolumn{1}{c|@{}}{\textbf{1.68}} 
                & \mc{\textbf{41.17}} & \mc{\textbf{80.21}}
            \\ 
            \bottomrule
        \end{tabular}
        }
    \end{threeparttable}
    \end{adjustbox}
    \caption{Comparison on the FMAVS dataset under $\pm 20 \text{m}$ position noise and $\pm 10^\circ$ orientation noise. \RGB, 4\RGB, \LiDAR, and \Lp\ denote 1 RGB camera, 4 RGB cameras, LiDAR, and LiDAR-pretrained model, respectively. $\dagger$ indicates reproduced results under revised settings from~\cite{wang2023satellite}.}
    \label{tab:shan}
    \end{centering}
\end{table*} 
\paragraph{KITTI dataset}
In Table~\ref{tab:shi}, we evaluated the proposed method against the recent cross-view localization methods on the cross-view KITTI dataset~\cite{geiger2013vision, shi2022beyond}. Our method achieved state-of-the-art results under a $\pm 20\text{m}$ and $\pm10^\circ$ initial pose error in both same-area and cross-area scenarios, demonstrating robustness across diverse conditions.

The proposed method directly optimizes pose, achieving a mean position error of $1.15\text{m}$ and a mean orientation error of $0.11^\circ$ in the same-area setting. In comparison, DSM~\cite{shi2020looking} and SliceMatch~\cite{lentsch2023slicematch} partition satellite-view features into patches to match ground-view features, which limits both the accuracy and precision of localization. 

LM optimizer-based cross-view pose optimization methods~\cite{wang2023satellite, shi2022beyond, wang2023view, wang2024view},  outperformed conventional cross-view feature matching methods by refining the pose. However, they solely rely on the \tcolor{given} pose, often converging to a local optimum in repetitive patterns. This limitation results in lower longitudinal performance. In contrast, our method integrates cross-view features at a given pose, global context, and feature difference gradients to capture distinctive features within repetitive patterns, achieving significantly higher longitudinal recall rates within $1\text{m}$ of $94.22\%$ in same-area and $50.02\%$ in cross-area settings, even with large initial pose errors.

Boosting~\cite{shi2023boosting} employs a two-stage framework combining pose optimization with dense feature matching to incorporate a global context. However, the independent optimization of each stage limits its ability to fully integrate the global context, leading to an orientation bias. In contrast, our unified DNN-based network integrates the global context with the PID branches, surpassing Boosting~\cite{shi2023boosting} and mitigating orientation bias. CCVPE~\cite{xia2023convolutional} and Song~\etal~\cite{song2024learning} improved performance with the highly expressive decoder and RAFT~\cite{teed2020raft}, respectively, but both struggled in cross-area settings. By contrast, our PIDLoc leverages global context and spatial relationships, achieving superior generalization performance.

\paragraph{Ford Multi-AV dataset}
In Table~\ref{tab:shan}, we evaluated our method under diverse weather and cross-area conditions using the FMAVS dataset. Consistent with SIBCL~\cite{wang2023satellite}, training was conducted using only log 4 data from FMAVS, employing data-efficient training to evaluate generalization performance. The evaluation was conducted on log 4 data for different weather conditions (same area) and log 5 data for different areas and weather conditions (cross area). 

The proposed method achieved recall rates of $41.92\%$, $29.03\%$, and $41.17\%$ in the lateral, longitudinal, and rotational directions, respectively, demonstrating robustness to different weather and cross-area settings. However, existing methods~\cite{wang2023satellite, wang2023view} neglect global context and spatial relationships among the features, resulting in significant performance drops in cross-area settings. SIBCL~\cite{wang2023satellite} and PureACL (1C)~\cite{wang2023view} show mean position errors of $4.19\text{m}$ and $5.45\text{m}$ in same-area settings, respectively. However, their performances dropped significantly in cross-area settings with errors of $15.20\text{m}$ and $6.50\text{m}$. PureACL (4C)~\cite{wang2023view} leverages four cameras to capture global context and spatial relationships across viewpoints, improving cross-area performance. However, the proposed method achieved superior performances even with a single camera by effectively capturing global context and learning spatial relationships between features.

\subsection{Ablation studies}

\begin{figure}[t!]
    \centering
    \includegraphics[width=1.0\columnwidth]{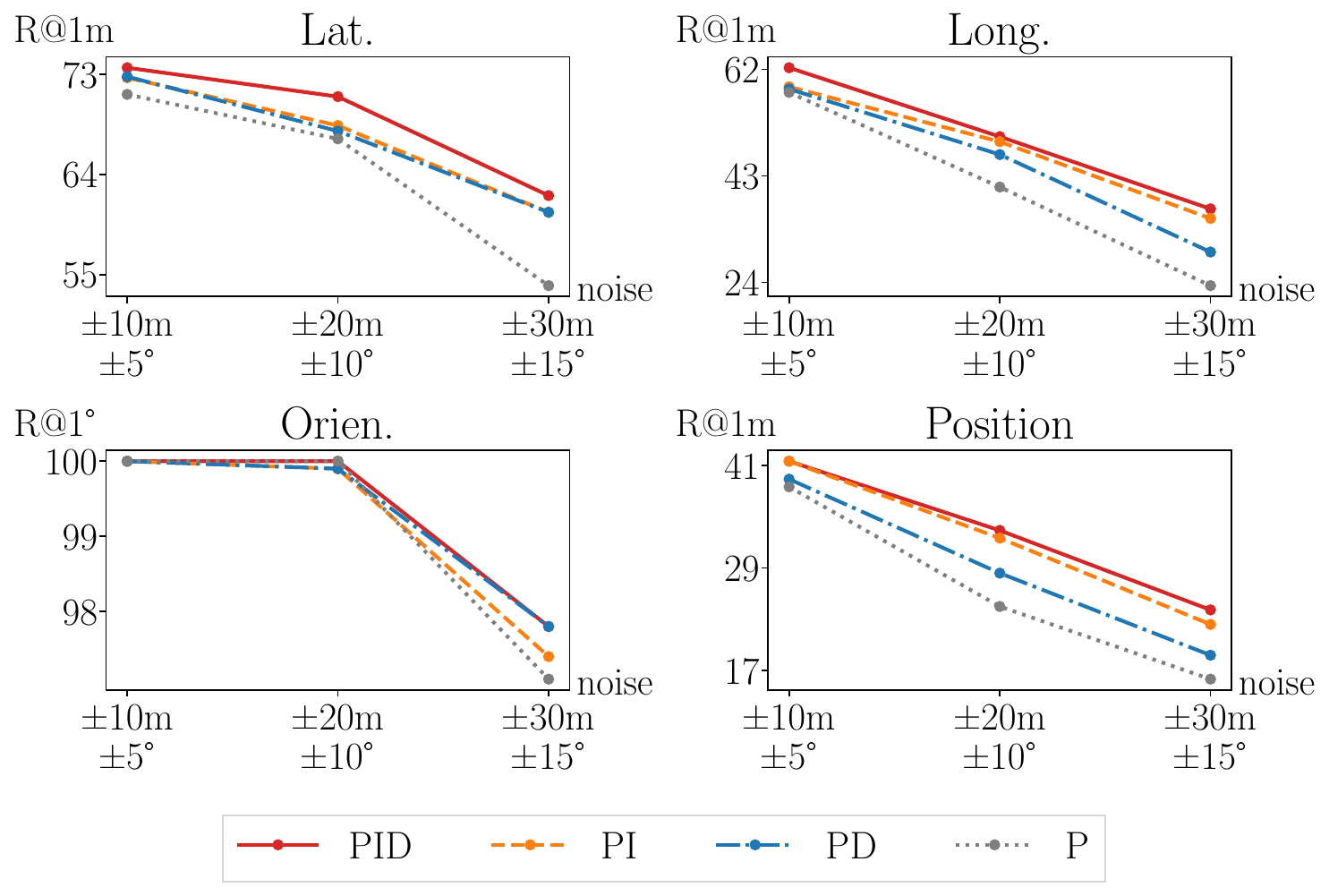}
    \caption{Ablation analysis of the PID branches under varying initial pose errors in the cross-area setting of the KITTI dataset.}
    \label{fig:varying}
\end{figure}
\paragraph{Robustness to the initial pose errors}
In Fig.~\ref{fig:varying}, we evaluated the localization performance under varying initial pose errors in the cross-area setting of the KITTI dataset. The P branch focuses solely on feature differences at the given pose, performing well under small initial pose errors of $\pm 10\text{m}$. However, the performance degraded substantially under larger pose errors of $\pm 20\text{m}$ and $\pm 30\text{m}$. This degradation is especially severe along the lateral and longitudinal directions, where repetitive patterns cause convergence to a local optimum due to the limited context. The I branch addresses this limitation by incorporating the global context with pose candidates, mitigating performance degradation under large initial pose errors. The D branch integrates fine-grained context through feature difference gradients, enabling precise pose estimation within $1\text{m}$ and improving overall performance. When combined with the P branch, the I and D branches improve performance in all directions, particularly under initial pose errors of $\pm 30\text{m}$.

\begin{table}[t!]
\begin{center}
    {\footnotesize 
        \setlength\tabcolsep{4pt}
        \begin{tabular}{@{} c *{8}{c}@{}}
            \toprule 
                \multicolumn{1}{c}{\multirow{2}{*}{Gradient}} & \multicolumn{1}{c}{Lat. (\%) ↑} & 
                \multicolumn{1}{c}{Long. (\%) ↑} &
                \multicolumn{1}{c}{Orien. (\%) ↑}  
            \\ 
                \multicolumn{1}{c}{} &
                \multicolumn{1}{c}{R@$0.25\text{m}$} &
                \multicolumn{1}{c}{R@$0.25\text{m}$} &
                \multicolumn{1}{c}{R@$0.25^{\circ}$} \\ 
            \midrule
                \mc{None} & 19.03 & 13.54 & 58.34\\ 
                \mc{Lat.} & 22.67 & 13.70 & 66.33 \\  
                \mc{Long.} & 21.47 & 14.24 & 64.11 \\   
                \mc{Orien.} & 20.19 & 14.53 & 69.52 \\ 
            \midrule
                \mc{All}  &  \mc{\textbf{24.06}}  & \mc{\textbf{15.13}}  &  \mc{\textbf{72.17}} \\ 
            \bottomrule 
        \end{tabular}
    }
\caption{Ablation analysis of the feature difference gradients in the D branch. The PI branches are used as the baseline.} \label{tab:feature_gradient}
\end{center}
\end{table}

\begin{table}[t!]
\begin{center}
    {\footnotesize 
        \setlength\tabcolsep{4pt}
        \begin{tabular}{@{} c *{8}{c}@{}}
            \toprule 
                \multicolumn{1}{c}{\multirow{2}{*}{Method}} & 
                \multicolumn{1}{c}{\multirow{2}{*}{PE}} & 
                \multicolumn{1}{c}{Lat. (\%) ↑} & 
                \multicolumn{1}{c}{Long. (\%) ↑} &
                \multicolumn{1}{c}{Orien. (\%) ↑}  
            \\ 
                \multicolumn{1}{c}{} & 
                \multicolumn{1}{c}{} & 
                \multicolumn{1}{c}{R@$1\text{m}$} &
                \multicolumn{1}{c}{R@$1\text{m}$} &
                \multicolumn{1}{c}{R@$1^{\circ}$} \\ 
            \midrule
                \mc{AvgPool feature} & \checkmark & 53.02 & 25.05 & 99.22\\
                \mc{WeightedAvg pose} & \checkmark & 59.94 & 32.99 & 97.30 \\
                \mc{MaxPool feature} & \checkmark & 62.62 & 41.55 & 90.11 \\
            \midrule
                \mc{SPE (Ours)} & \textsf{X} & 66.06 & 42.37 & 99.88 \\
                \mc{\textbf{\dnnpose{} (Ours)}} & \checkmark &  \mc{\textbf{71.01}}  & \mc{\textbf{50.02}}  &  \mc{\textbf{99.96}} \\
            \bottomrule 
        \end{tabular}
    }
\caption{Ablation analysis of the spatial relationship modeling.} \label{tab:dnn_pose}
\end{center}
\end{table}
\paragraph{Impact of feature difference gradients on precision}
In Table~\ref{tab:feature_gradient}, we evaluated the impact of the feature difference gradients for lateral, longitudinal, and rotational directions on localization precision in the D branch, using the PI branches as the ablation baseline.
The precision was measured using the recall rates within $0.25\text{m}$ and $0.25^{\circ}$ under initial pose errors of $\pm 20\text{m}$ and $\pm 10^{\circ}$. As shown in Table~\ref{tab:feature_gradient}, utilizing feature difference gradients in each direction improved localization precision not only in the respective direction but also across all the directions. This demonstrates the interdependence of lateral, longitudinal, and rotational components during pose refinement. By combining feature difference gradients from all directions, the proposed D branch improved localization precision, with recall rates of $24.06\%$, $15.13\%$, and $72.17\%$.

\paragraph{Spatial relationship modeling}
Table~\ref{tab:dnn_pose} shows the ablation analysis of the pose estimator methods in the cross-area setting of the KITTI dataset, using the PID-branch features and positional embeddings as input. The proposed \dnnpose{} \tcolor{with positional embedding (PE)} explicitly models the spatial relationships among local features, achieving the highest recall rates of $71.01\%$, $50.02\%$, and $99.96\%$ for the lateral, longitudinal, and rotational directions, respectively. \tcolor{PE further enhanced \dnnpose{} by encoding position-dependent representations, which are crucial for accurate pose estimation.}

By contrast, the average pooling method~\cite{shi2023boosting} neglects local features, resulting in low localization performance. Weighted averaging pose methods~\cite{shi2022beyond, wang2023satellite, wang2023view} estimate the pose for each feature and average them based on confidence scores. However, they fail to capture spatial relationships among features, converging to a local optimum in the lateral and longitudinal directions. 
The max pooling method improved localization performance but neglected the global feature, reducing orientation performance. In contrast, the proposed \dnnpose{} captures spatial interactions among local features, effectively integrating global context to enhance position and orientation estimation accuracy.

\subsection{Qualitative analysis}
\begin{figure}[t!]
    \centering
    \begin{subfigure}{0.47\columnwidth}
        \centering
        \includegraphics[width=\linewidth]{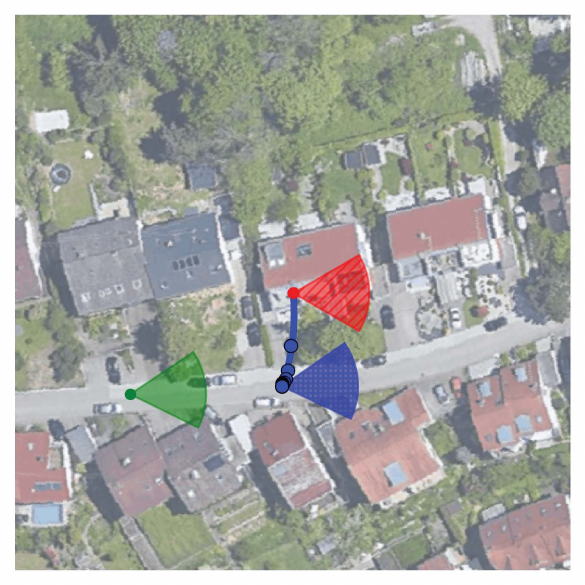} 
        \caption{SIBCL~\cite{wang2023satellite}}
        \label{fig:subfig_a}
    \end{subfigure}
    \hfill
    \begin{subfigure}{0.47\columnwidth}
        \centering
        \includegraphics[width=\linewidth]{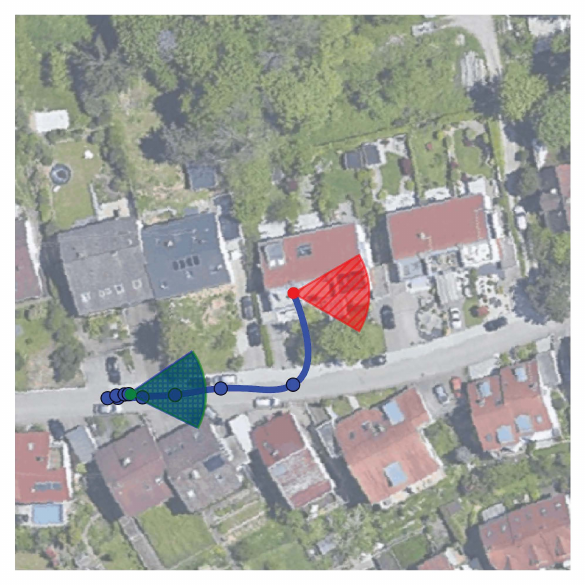} 
        \caption{PIDLoc (Ours)}
        \label{fig:subfig_b}
    \end{subfigure}
    \caption{Visualization of localization results. The red, green, and blue circular sectors represent the current, ground-truth, and predicted pose, respectively. The blue line represents the iterative trajectory of predicted poses. Compared with SIBCL~\cite{wang2023satellite}, PIDLoc converges to the global optimum in a challenging environment with repetitive patterns.}
    \label{fig:qualitative}
\end{figure}

Fig.~\ref{fig:qualitative} compares the SIBCL~\cite{wang2023satellite} with our PIDLoc under an initial pose error of $\pm{30}\text{m}$ and $\pm{15}^\circ$ on the KITTI dataset. The SIBCL~\cite{wang2023satellite} converges to a local optimum in repetitive patterns, such as building facades and vegetation rows, due to its limited FoV. In contrast, PIDLoc leverages global and fine-grained contexts, converging to the global optimum. Additional results in the supplementary material highlight PIDLoc's robustness under large initial pose errors.

\section{Conclusion}
This paper proposes PIDLoc, a novel cross-view pose optimization network inspired by the PID controller. The PIDLoc leverages the local, global, and fine-grained contexts to enhance localization accuracy under large initial pose errors. Extensive experiments on the KITTI and FMAVS datasets demonstrate the superiority and validity of our method. In future work, we will integrate the proposed PIDLoc with a simultaneous localization and mapping (SLAM) pipeline to eliminate loop closure dependency and improve initial pose accuracy for localization.

%% file: main.bbl
\begin{thebibliography}{34}
\providecommand{\natexlab}[1]{#1}
\providecommand{\url}[1]{\texttt{#1}}
\expandafter\ifx\csname urlstyle\endcsname\relax
  \providecommand{\doi}[1]{doi: #1}\else
  \providecommand{\doi}{doi: \begingroup \urlstyle{rm}\Url}\fi

\bibitem[Agarwal et~al.(2020)Agarwal, Vora, Pandey, Williams, Kourous, and McBride]{agarwal2020ford}
Siddharth Agarwal, Ankit Vora, Gaurav Pandey, Wayne Williams, Helen Kourous, and James McBride.
\newblock {Ford multi-AV seasonal dataset}.
\newblock \emph{The International Journal of Robotics Research}, 39\penalty0 (12):\penalty0 1367--1376, 2020.

\bibitem[Choi and Myung(2020)]{choi2020brm}
Junho Choi and Hyun Myung.
\newblock {BRM localization: UAV localization in GNSS-denied environments based on matching of numerical map and UAV images}.
\newblock In \emph{IEEE/RSJ International Conference on Intelligent Robots and Systems}, pages 4537--4544, 2020.

\bibitem[Deng et~al.(2009)Deng, Dong, Socher, Li, Li, and Fei-Fei]{deng2009imagenet}
Jia Deng, Wei Dong, Richard Socher, Li-Jia Li, Kai Li, and Li Fei-Fei.
\newblock {ImageNet: A large-scale hierarchical image database}.
\newblock In \emph{Proceedings of the IEEE/CVF Conference on Computer Vision and Pattern Recognition}, pages 248--255, 2009.

\bibitem[Deuser et~al.(2023)Deuser, Habel, and Oswald]{deuser2023sample4geo}
Fabian Deuser, Konrad Habel, and Norbert Oswald.
\newblock {Sample4Geo: Hard negative sampling for cross-view geo-localisation}.
\newblock In \emph{Proceedings of the IEEE/CVF International Conference on Computer Vision}, pages 16847--16856, 2023.

\bibitem[Fervers et~al.(2022)Fervers, Bullinger, Bodensteiner, Arens, and Stiefelhagen]{fervers2022continuous}
Florian Fervers, Sebastian Bullinger, Christoph Bodensteiner, Michael Arens, and Rainer Stiefelhagen.
\newblock {Continuous self-localization on aerial images using visual and LiDAR sensors}.
\newblock In \emph{IEEE/RSJ International Conference on Intelligent Robots and Systems}, pages 7028--7035, 2022.

\bibitem[Franklin et~al.(2002)Franklin, Powell, Emami-Naeini, and Powell]{franklin2002feedback}
Gene~F Franklin, J~David Powell, Abbas Emami-Naeini, and J~David Powell.
\newblock \emph{{Feedback Control of Dynamic Systems}}.
\newblock Prentice Hall Upper Saddle River, 2002.

\bibitem[Geiger et~al.(2013)Geiger, Lenz, Stiller, and Urtasun]{geiger2013vision}
Andreas Geiger, Philip Lenz, Christoph Stiller, and Raquel Urtasun.
\newblock {Vision meets robotics: The KITTI dataset}.
\newblock \emph{The International Journal of Robotics Research}, 32\penalty0 (11):\penalty0 1231--1237, 2013.

\bibitem[Hu et~al.(2018)Hu, Feng, Nguyen, and Lee]{hu2018cvm}
Sixing Hu, Mengdan Feng, Rang~MH Nguyen, and Gim~Hee Lee.
\newblock {CVM-Net: Cross-view matching network for image-based ground-to-aerial geo-localization}.
\newblock In \emph{Proceedings of the IEEE/CVF Conference on Computer Vision and Pattern Recognition}, pages 7258--7267, 2018.

\bibitem[Kingma and Ba(2015)]{KingBa15}
Diederik Kingma and Jimmy Ba.
\newblock {ADAM: A method for stochastic optimization}.
\newblock In \emph{International Conference on Learning Representations}, 2015.

\bibitem[Lee et~al.(2023)Lee, Song, Lim, Lee, and Myung]{lee20232}
Alex~Junho Lee, Seungwon Song, Hyungtae Lim, Woojoo Lee, and Hyun Myung.
\newblock {(LC)$^2$: LiDAR-camera loop constraints for cross-modal place recognition}.
\newblock \emph{IEEE Robotics and Automation Letters}, 8\penalty0 (6):\penalty0 3589--3596, 2023.

\bibitem[Lentsch et~al.(2023)Lentsch, Xia, Caesar, and Kooij]{lentsch2023slicematch}
Ted Lentsch, Zimin Xia, Holger Caesar, and Julian~FP Kooij.
\newblock {SliceMatch: Geometry-guided aggregation for cross-view pose estimation}.
\newblock In \emph{Proceedings of the IEEE/CVF Conference on Computer Vision and Pattern Recognition}, pages 17225--17234, 2023.

\bibitem[Liu and Li(2019)]{liu2019lending}
Liu Liu and Hongdong Li.
\newblock {Lending orientation to neural networks for cross-view geo-localization}.
\newblock In \emph{Proceedings of the IEEE/CVF Conference on Computer Vision and Pattern Recognition}, pages 5624--5633, 2019.

\bibitem[Mor{\'e}(2006)]{more2006levenberg}
Jorge~J Mor{\'e}.
\newblock {The Levenberg-Marquardt algorithm: Implementation and theory}.
\newblock In \emph{Numerical Analysis: Proceedings of the Biennial Conference}, pages 105--116. Springer, 2006.

\bibitem[Pirayesh and Zeng(2022)]{pirayesh2022jamming}
Hossein Pirayesh and Huacheng Zeng.
\newblock {Jamming attacks and anti-jamming strategies in wireless networks: A comprehensive survey}.
\newblock \emph{IEEE Communications Surveys \& Tutorials}, 24\penalty0 (2):\penalty0 767--809, 2022.

\bibitem[Reid et~al.(2019)Reid, Houts, Cammarata, Mills, Agarwal, Vora, and Pandey]{reid2019localization}
Tyler~GR Reid, Sarah~E Houts, Robert Cammarata, Graham Mills, Siddharth Agarwal, Ankit Vora, and Gaurav Pandey.
\newblock Localization requirements for autonomous vehicles.
\newblock \emph{SAE International Journal of Connected and Automated Vehicles}, 2\penalty0 (3), 2019.

\bibitem[Sarlin et~al.(2023)Sarlin, DeTone, Yang, Avetisyan, Straub, Malisiewicz, Bulo, Newcombe, Kontschieder, and Balntas]{sarlin2023orienternet}
Paul-Edouard Sarlin, Daniel DeTone, Tsun-Yi Yang, Armen Avetisyan, Julian Straub, Tomasz Malisiewicz, Samuel~Rota Bulo, Richard Newcombe, Peter Kontschieder, and Vasileios Balntas.
\newblock {OrienterNet: Visual localization in 2D public maps with neural matching}.
\newblock In \emph{Proceedings of the IEEE/CVF Conference on Computer Vision and Pattern Recognition}, pages 21632--21642, 2023.

\bibitem[Shi and Li(2022)]{shi2022beyond}
Yujiao Shi and Hongdong Li.
\newblock {Beyond cross-view image retrieval: Highly accurate vehicle localization using satellite image}.
\newblock In \emph{Proceedings of the IEEE/CVF Conference on Computer Vision and Pattern Recognition}, pages 17010--17020, 2022.

\bibitem[Shi et~al.(2020{\natexlab{a}})Shi, Yu, Campbell, and Li]{shi2020looking}
Yujiao Shi, Xin Yu, Dylan Campbell, and Hongdong Li.
\newblock {Where am I looking at? Joint location and orientation estimation by cross-view matching}.
\newblock In \emph{Proceedings of the IEEE/CVF Conference on Computer Vision and Pattern Recognition}, pages 4064--4072, 2020{\natexlab{a}}.

\bibitem[Shi et~al.(2020{\natexlab{b}})Shi, Yu, Liu, Zhang, and Li]{shi2020optimal}
Yujiao Shi, Xin Yu, Liu Liu, Tong Zhang, and Hongdong Li.
\newblock {Optimal feature transport for cross-view image geo-localization}.
\newblock In \emph{Proceedings of the AAAI Conference on Artificial Intelligence}, pages 11990--11997, 2020{\natexlab{b}}.

\bibitem[Shi et~al.(2023)Shi, Wu, Perincherry, Vora, and Li]{shi2023boosting}
Yujiao Shi, Fei Wu, Akhil Perincherry, Ankit Vora, and Hongdong Li.
\newblock {Boosting 3-DoF ground-to-satellite camera localization accuracy via geometry-guided cross-view transformer}.
\newblock In \emph{Proceedings of the IEEE/CVF International Conference on Computer Vision}, pages 21516--21526, 2023.

\bibitem[Shi et~al.(2025)Shi, Li, Perincherry, and Vora]{shi2025weakly}
Yujiao Shi, Hongdong Li, Akhil Perincherry, and Ankit Vora.
\newblock {Weakly-supervised camera Localization by ground-to-satellite image registration}.
\newblock In \emph{European Conference on Computer Vision}, pages 39--57, 2025.

\bibitem[Simonyan and Zisserman(2014)]{simonyan2014very}
Karen Simonyan and Andrew Zisserman.
\newblock {Very deep convolutional networks for large-scale image recognition}.
\newblock \emph{arXiv preprint arXiv:1409.1556}, 2014.

\bibitem[Song et~al.(2024)Song, Lu, Shi, et~al.]{song2024learning}
Zhenbo Song, Jianfeng Lu, Yujiao Shi, et~al.
\newblock Learning dense flow field for highly-accurate cross-view camera localization.
\newblock \emph{Advances in Neural Information Processing Systems}, 36, 2024.

\bibitem[Teed and Deng(2020)]{teed2020raft}
Zachary Teed and Jia Deng.
\newblock {RAFT: Recurrent all-pairs field transforms for optical flow}.
\newblock In \emph{European Conference on Computer Vision}, pages 402--419, 2020.

\bibitem[Toker et~al.(2021)Toker, Zhou, Maximov, and Leal-Taix{\'e}]{toker2021coming}
Aysim Toker, Qunjie Zhou, Maxim Maximov, and Laura Leal-Taix{\'e}.
\newblock {Coming down to earth: Satellite-to-street view synthesis for geo-localization}.
\newblock In \emph{Proceedings of the IEEE/CVF Conference on Computer Vision and Pattern Recognition}, pages 6488--6497, 2021.

\bibitem[Van~Diggelen and Enge(2015)]{van2015world}
Frank Van~Diggelen and Per Enge.
\newblock {The world’s first GPS MOOC and worldwide laboratory using smartphones}.
\newblock In \emph{Proceedings of the International Technical Meeting of the Satellite Division of the Institute of Navigation}, pages 361--369, 2015.

\bibitem[Wang et~al.(2023{\natexlab{a}})Wang, Zhang, Perincherry, Vora, and Li]{wang2023view}
Shan Wang, Yanhao Zhang, Akhil Perincherry, Ankit Vora, and Hongdong Li.
\newblock {View consistent purification for accurate cross-view localization}.
\newblock In \emph{Proceedings of the IEEE/CVF International Conference on Computer Vision}, pages 8197--8206, 2023{\natexlab{a}}.

\bibitem[Wang et~al.(2023{\natexlab{b}})Wang, Zhang, Vora, Perincherry, and Li]{wang2023satellite}
Shan Wang, Yanhao Zhang, Ankit Vora, Akhil Perincherry, and Hongdong Li.
\newblock {Satellite image based cross-view localization for autonomous vehicle}.
\newblock In \emph{IEEE International Conference on Robotics and Automation}, pages 3592--3599, 2023{\natexlab{b}}.

\bibitem[Wang et~al.(2024{\natexlab{a}})Wang, Nguyen, Liu, Zhang, Muthu, Maken, Zhang, and Li]{wang2024view}
Shan Wang, Chuong Nguyen, Jiawei Liu, Yanhao Zhang, Sundaram Muthu, Fahira~Afzal Maken, Kaihao Zhang, and Hongdong Li.
\newblock {View from above: Orthogonal-view aware cross-view localization}.
\newblock In \emph{Proceedings of the IEEE/CVF Conference on Computer Vision and Pattern Recognition}, pages 14843--14852, 2024{\natexlab{a}}.

\bibitem[Wang et~al.(2024{\natexlab{b}})Wang, Xu, Cui, Wan, and Zhang]{wang2024fine}
Xiaolong Wang, Runsen Xu, Zhuofan Cui, Zeyu Wan, and Yu Zhang.
\newblock {Fine-grained cross-view geo-localization using a correlation-aware homography estimator}.
\newblock \emph{Advances in Neural Information Processing Systems}, 36, 2024{\natexlab{b}}.

\bibitem[Xia et~al.(2022)Xia, Booij, Manfredi, and Kooij]{xia2022visual}
Zimin Xia, Olaf Booij, Marco Manfredi, and Julian~FP Kooij.
\newblock {Visual cross-view metric localization with dense uncertainty estimates}.
\newblock In \emph{European Conference on Computer Vision}, pages 90--106, 2022.

\bibitem[Xia et~al.(2023)Xia, Booij, and Kooij]{xia2023convolutional}
Zimin Xia, Olaf Booij, and Julian~FP Kooij.
\newblock {Convolutional cross-view pose estimation}.
\newblock \emph{IEEE Transactions on Pattern Analysis and Machine Intelligence}, 2023.

\bibitem[Zhu et~al.(2021)Zhu, Yang, and Chen]{zhu2021vigor}
Sijie Zhu, Taojiannan Yang, and Chen Chen.
\newblock {VIGPR: Cross-view image geo-localization beyond one-to-one retrieval}.
\newblock In \emph{Proceedings of the IEEE/CVF Conference on Computer Vision and Pattern Recognition}, pages 3640--3649, 2021.

\bibitem[Zidan et~al.(2020)Zidan, Adegoke, Kampert, Birrell, Ford, and Higgins]{zidan2020gnss}
Jasmine Zidan, Elijah~I Adegoke, Erik Kampert, Stewart~A Birrell, Col~R Ford, and Matthew~D Higgins.
\newblock {GNSS vulnerabilities and existing solutions: A review of the literature}.
\newblock \emph{IEEE Access}, 9:\penalty0 153960--153976, 2020.

\end{thebibliography}
